\documentclass[a4paper,twocolumn,australian,english,conference]{IEEEtran}
\ifCLASSINFOpdf
  \usepackage[pdftex]{graphicx}
  \DeclareGraphicsExtensions{.pdf,.jpeg,.png}
\else
  \usepackage[dvips]{graphicx}
  \DeclareGraphicsExtensions{.eps}
\fi
\usepackage[cmex10]{amsmath}
\usepackage{amssymb}

\usepackage{tikz}
\usepackage{textcomp}
\usepackage{hyperref}

\newcommand\copyrighttext{%
  \footnotesize \textcopyright 2016 IEEE. Personal use of this material is permitted.
  Permission from IEEE must be obtained for all other uses, in any current or future
  media, including reprinting/republishing this material for advertising or promotional
  purposes, creating new collective works, for resale or redistribution to servers or
  lists, or reuse of any copyrighted component of this work in other works.
  DOI: \href{<http://ieeexplore.ieee.org>}{<DOI No. TBD>}}
\newcommand\copyrightnotice{%
\begin{tikzpicture}[remember picture,overlay]
\node[anchor=south,yshift=10pt] at (current page.south) {\fbox{\parbox{\dimexpr\textwidth-\fboxsep-\fboxrule\relax}{\copyrighttext}}};
\end{tikzpicture}%
}

\begin{document}

\title{Understanding data augmentation for classification: when to warp?}

\author{\IEEEauthorblockN{Sebastien C. Wong}\IEEEauthorblockA{Defence Science and Technology\\
Edinburgh \\
SA, Australia\\
Email: sebastien.wong@dsto.defence.gov.au}\and \IEEEauthorblockN{Adam Gatt}\IEEEauthorblockA{Australian Defence Force\\
Edinburgh \\
SA, Australia}\and \IEEEauthorblockN{Victor Stamatescu\\
and Mark D. McDonnell}\IEEEauthorblockA{Computational Learning Systems Laboratory\\
Information Technology and Mathematical Sciences\\
University of South Australia\\
Mawson Lakes\\
SA, Australia}}
\maketitle
\copyrightnotice

\begin{abstract}
In this paper we investigate the benefit of augmenting data with synthetically
created samples when training a machine learning classifier. Two approaches
for creating additional training samples are \emph{data warping},
which generates additional samples through transformations applied
in the data-space, and \emph{synthetic over-sampling}, which creates
additional samples in feature-space. We experimentally evaluate the
benefits of data augmentation for a convolutional backpropagation-trained
neural network, a convolutional support vector machine and a convolutional
extreme learning machine classifier, using the standard MNIST handwritten
digit dataset. We found that while it is possible to perform generic
augmentation in feature-space, if plausible transforms for the data
are known then augmentation in data-space provides a greater benefit
for improving performance and reducing overfitting. 
\end{abstract}

\IEEEpeerreviewmaketitle{}

\section{Introduction}

The competition winning classifiers described in \cite{krizhevsky2012imagenet,simonyan2014very,szegedy2015going,he2015deep}
all utilized techniques to artificially increase the number of training
examples. Previous research to systematically understand the benefits
and limitations of data augmentation demonstrate that data augmentation
can act as a regularizer in preventing overfitting in neural networks
\cite{simard2003best,ciresan2010deep} and improve performance in
imbalanced class problems \cite{chawla2002smote}. In this paper we
empirically investigate the benefits and limitations of data augmentation
on three machine learning classifiers, and attempt to answer the question
of \emph{how and when to apply data augmentation?}

To enable comparison of data augmentation for a convolutional neural
network (CNN), a convolutional support vector machine (CSVM) \cite{coates2011analysis},
and a convolutional extreme learning machine (CELM) \cite{mcdonnell2015enhanced}
classifier, we use the well known MNIST handwritten digit dataset
\cite{lecun1998gradient}.

\section{Previous Work}

This section provides a brief review of previous works that have used
augmented training data to improve classifier performance. 

The concept of \emph{data warping} (for neural networks) can possibly
be attributed to \cite{baird1992document} in creating a model of
character distortions that could occur in the printing (or handwriting)
process, the model described deformations and defects needed to generate
synthetic character examples. The term \emph{warping} was coined by
\cite{yaeger1996effective}, in the context of producing randomly
warped stroke data for handwriting character classification. The warped
character stroke data was used to balance the amount of training examples
for each character class in order to reduce the bias in the classifier
to favor more frequently presented training examples. This approach
was extended by \cite{simard2003best} to improve the performance
of a standard backpropagation-trained neural network, and achieve
record performance (in 2003) of 0.4\% error rate on the MNIST handwritten
digit database using a convolutional neural network. The warped training
data was created by applying both affine transformations (translation,
shearing, rotation) and elastic distortions to images of existing
character examples. The elastic distortions were generated using a
scaled normalized random displacement field in image space, and are
stated to correspond to uncontrolled oscillations of the hand muscles.
This approach was further extended to create a simple deep neural
network that was only trained on warped data \cite{ciresan2010deep},
where new warped training data was created for each epoch of the backpropagation
algorithm. This allowed for a neural network with a large number of
parameters that overcame the issues of over-fitting and diminished
convergence by the backpropagation algorithm during training, which
achieved a state of the art (in 2010) error rate of 0.35\% on the
MNIST database.

The problem of class imbalance, where real-world datasets often only
contain a small percentage of ``interesting'' or target class examples,
was addressed by the use of a Synthetic Minority Over-Sampling Technique
(SMOTE) \cite{chawla2002smote}. In SMOTE the synthetic examples are
created in feature-space from randomly selected pairs of real world
feature-examples from the minority class. The advantage of \emph{synthetic
over-sampling} compared to the \emph{data warping} approach, is that
synthetic examples are created in feature-space, and thus the SMOTE
algorithm is application independent.

Both the SMOTE and data warping approaches modify real world examples
to create augmented data. An alternative would be to generate synthetic
imagery. This approach was investigated by \cite{zhang2015learning},
in creating synthetic images of building roofs to augment real image
data. By visualizing the data in feature-space they found that the
distributions of artificially generated images and real images (that
were the basis of the synthetic examples) were displaced. The authors
termed this problem the \emph{synthetic gap}. To overcome this issue
they trained a sparse auto-encoder simultaneously with real and synthetic
images, such that the learnt features minimized the synthetic gap. 

The issue of the difference between real and synthetic examples is
discussed by \cite{sato2015apac}, in using a classifier that was
trained with augmented data only. The paper argues that to maximize
the probability of correct classification any test sample should undergo
the same warping process as the training samples.

\section{Method and Data}

\begin{figure}[t]
\includegraphics[width=0.99\columnwidth]{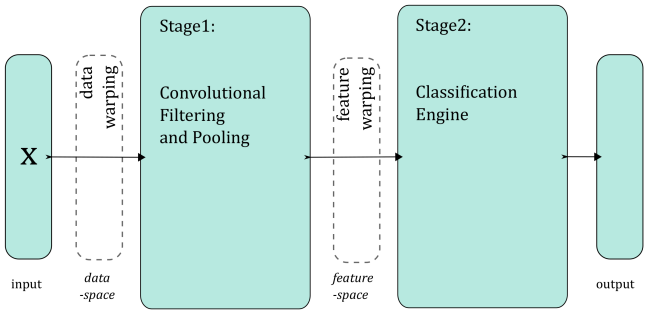}

\caption{Experimental architecture. Stage 1 was kept the same for all experiments,
while stage 2 consisted of either a backpropagation-trained neural
network, support vector machine or extreme learning machine classifier. }
\label{figure_one}
\end{figure}

The literature suggests that it is possible to augment training data
in \emph{data-space} or \emph{feature-space}. To understand the benefits
of each approach the experimental architecture (for each machine learning
classifier) was structured into two stages, such that there was a
clear distinction between data-space and feature-space, illustrated
in Figure \ref{figure_one}. Each system under evaluation had the
same convolutional and pooling layer that was used to generate features
for the classification engine. The weights for this convolutional
and pooling layer were kept fixed to ensure that the same features
were used for all experiments with the same dataset.

\subsection{Datasets}

The MNIST database \cite{lecun1998gradient} was used for the experiments.
This database consists of a training set of 60,000 labeled 28 by 28
pixel grayscale images of handwritten digits and a test set of 10,000
labelled images. There are 10 classes, the digits 0 to 9. 

An important question to be answered is how does using synthetic data
augmentation compare with collecting more real data? To answer this
question a set of baseline performance figures for each of the classifiers
were created by reducing the amount of data available in training.
The quantity of samples from each class was kept equal to remove any
impact from class imbalance, which reduced the total number of available
training samples to 50,000. These results can then be compared with
the performance of adding an equivalent amount of augmented data.
Here the augmented data was generated from a very small pool of 500
real training samples from each class.

\subsection{Augmentation in data-space}

For image data it is possible to create plausible transformations
of existing samples that preserves label information, with the validation
of label integrity being performed by a human observer (can a human
still recognize the object). One of the significant improvements in
performance of classifiers on the MNIST database was through the introduction
of elastic deformations \cite{simard2003best}, in addition to the
existing affine transformations, for data augmentation. The elastic
deformation was performed by defining a normalized random displacement
field $\boldsymbol{u}(x,y)$ that for each pixel location $(x,y)$
in an image specifies a unit displacement vector, such that $\boldsymbol{R_{w}}=\boldsymbol{R_{o}}+\boldsymbol{\mathrm{\alpha\boldsymbol{u}}}$,
where $\boldsymbol{R_{w}}$ and $\boldsymbol{R_{o}}$ describe the
location of the pixels in the original and warped images respectively.
The strength of the displacement in pixels is given by $\boldsymbol{\alpha}$.
The smoothness of the displacement field is controlled by the parameter
$\sigma$, which is the standard deviation of the Gaussian that is
convolved with matrices of uniformly distributed random values that
form the \emph{$x$} and \emph{$y$} dimensions of the displacement
field $\boldsymbol{u}$. 

\begin{figure}[t]
\centering\includegraphics[width=0.8\columnwidth]{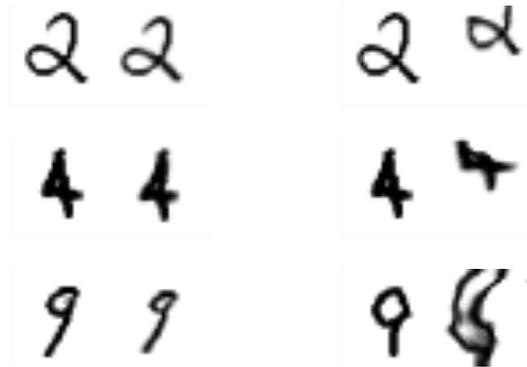}

\caption{Data warping using elastic deformations \cite{simard2003best}. Original MNIST digits
compared to warped digits for $\alpha=1.2$ (left) and $\alpha=8$
pixels (right).}
\label{figure_warpedimages}
\end{figure}

For the MNIST data set we found that large deformations, with the
displacement $\alpha\geq8$ pixels, could occasionally result in characters
that were difficult to recognize by a human observer, that is label
information was not preserved. This loss of label integrity was caused
by shifting a critical part of the character outside of image boundary,
or by introducing a ``kink'' that rendered the character illegible,
as illustrated in Figure \ref{figure_warpedimages}. We empirically
set $\alpha=1.2$ pixels with $\sigma=20$ based on performance of
the (quick to train) CELM algorithm. 

The samples needed for elastic warping were generated off-line and
the same data applied to each of the experiments.

\subsection{Augmentation in feature-space }

For some machine learning problems it may not be easy to validate
that an arbitrary transformation of an existing raw data sample has
indeed preserved label information. For these problems the features
may have been hand crafted to extract salient information on which
the classifier performs learning. The Synthetic Minority Over-Sampling
Technique (SMOTE) \cite{chawla2002smote} was inspired by the use
of data-warping to reduce class imbalance in the handwritten digit
problem. But by being applied in feature-space it was proposed by
the authors to be domain independent, and has since been used extensively
in medical research where datasets with a significant minority class
are common. For SMOTE a new synthetic sample is created by selecting
a random point in feature-space along a line intersecting \emph{k}
randomly chosen samples of the same class.

A more recent derivative of SMOTE is the Density Based SMOTE (DBSMOTE)
algorithm \cite{bunkhumpornpat2012dbsmote}, which generates the new
synthetic samples within a distance $eps$ of the cluster centre for
the class (following original notation). As the DBSMOTE algorithm
generates its synthetic samples around the center of each class, it
could be expected that these synthetic samples may not contribute
to reducing classifier overfitting. Indeed we found that using DBSMOTE
actually increased the amount of overfitting. 

For the MNIST data set we used $k=2$ random samples, for both SMOTE
and DBSMOTE, and distance threshold $eps\leq4$ for DBSMOTE.

The samples needed for SMOTE and DBSMOTE were generated online, as
these algorithms are naturally embedded within the classification
architecture, which is described in the next section.

\section{Experimental System Parameters}

As previously stated, the two-stage architecture illustrated in Figure
\ref{figure_one} was used for each experiment.

\subsection{Stage-1 convolution and pooling. }

For stage-1, to create a local receptive field, we adopt the method
and parameters described in \cite{mcdonnell2015enhanced}. Conceptually,
the $28\times28$ input image is convolved with a $W\times W$ filter,
where $W=7$ and the stride length is 1. This results in an intermediate
output feature layer with dimensions $(28-W-1)^{2}$. Next LP-pooling
is applied with a pool size of $q\times q$, where $q=8$, which results
in a down-sampled feature layer with dimensions $22\times22$. This
is repeated for $L$ filters, where $L=96$ and corresponds to the
first layer of pre-trained OverFeat filters \cite{sermanet2013overfeat}.

\subsection{Stage-2 classification.}

For stage-2, we evaluated the following classification schemes: backpropagation-trained
neural network, support vector machine, and extreme learning machine.
The weights of the convolutional layer were held constant.

The neural network classifier was a standard multi-layer perceptron
neural network with sigmoidal activation function. For simplicity
of parameters, and to keep the neural architecture similar between
classifiers, a single hidden layer with 1600 neurons was used. Learning
was performed using backpropagation on the the hidden neurons only
(and no changes were made to the convolutional layer). The number
of epochs (one batch application of backpropagation on all training
samples) was fixed at 2000. 

The multi-class support vector machine classifier, which used the
1-vs-all L2 loss function, was based on the code and parameters settings
of \cite{coates2011analysis}.

The extreme learning machine classifier, which used least squares
regression of random weight projection neurons, was based on the code
and parameter settings of \cite{mcdonnell2015enhanced}. For consistency
with the CNN, a single hidden layer with 1600 neurons was used.

\section{Experiments and Results}

The first step in understanding the benefit of data augmentation was
to vary the amount of real data samples available to train a classifier.
Our hypothesis is that for a given number of samples (of either real
or synthetic data) the benefit provided by real samples provides an
upper-bound that the performance of a classifier can be improved by
an equivalent number of synthetic samples. We denote this experiment
as our baseline, where \emph{training error} indicates the percentage
error rate (\emph{error \%}) of the classifier when evaluated against
the samples used to train the classifier, and \emph{test error} is
the \emph{error \%} on the predefined MNIST test set of 10,000 samples.

Each experiment is repeated three times to account for random variation.
The primary variation in the CNN is the initialization of the weights
and the order in which the training samples are presented to the classifier.
The variation in the CELM output is due to the random weight projection.
The CSVM implementation is deterministic, leading to no performance
variation for multiple runs.

\subsection{Baseline Results}

\begin{figure}[t]
\includegraphics[clip,width=0.99\columnwidth]{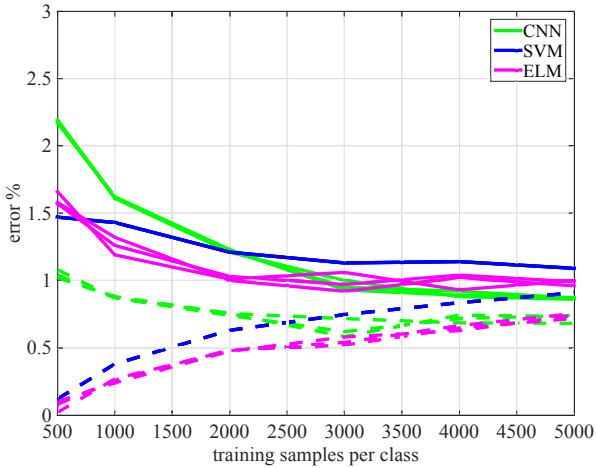}\caption{Baseline performance of CNN, CSVM and CELM on MNIST using real data.
The dashed lines indicate training error \%, and the solid lines indicate
test error \%. Lower test error \% indicates better performance. Reducing
the difference between training and test error \% is indicative of
reducing overfitting.}
\label{figure_two}
\end{figure}

The performance of the baseline system, illustrating the relative
performance of the CNN, SVM and ELM is shown in Figure \ref{figure_two}.
Each of the classifiers have similar test \emph{error \%} for the
amount of training data used. The primary observation is that as the
number of real samples are increased (from 500 to 5000 per class)
the gap between \emph{training error} and \emph{test error} decreases,
which indicates a reduction in overfitting. The amount of improvement
in performance varies across the classifiers. This confirms that the
primary benefit of increasing the number of samples is to reduce overfitting
(given that all other classifier parameters are held constant). 

Another interesting observation is that the CNN benefits the most
from additional training samples, in that increasing the number of
samples improving both the \emph{training error} and the \emph{test
error}. The increase in the amount of data results in more iterations
of the backpropagation algorithm for the same number of epochs. The
CELM varies the most in \emph{error \%} for a given amount of data,
due to the random weight projection neurons in the hidden layer. The
CELM also has good performance for small amounts of training data.
The CSVM also exhibited similar behaviour
to the CELM, with the best test \emph{error \%} for 500 samples per
class, but performance did not improve as much as the other classifiers
as the number of samples were increased.

\subsection{CNN Results}

\begin{figure}[t]
\includegraphics[width=0.99\columnwidth]{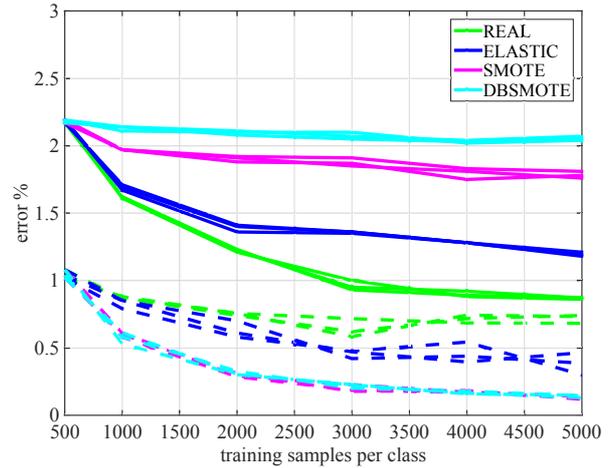}

\caption{CNN performance on MNIST. The dashed lines indicate training error
\%, and the solid lines indicate test error \%.}
\label{figure_three}
\end{figure}

The performance of the CNN is shown in Figure \ref{figure_three},
which illustrates how \emph{error \%} varies as the number of samples
are increased. Again this shows that increasing the number of samples
resulted in a performance improvement. Most notably the \emph{test
error \%} decreased steadily. The CNN results were consistent, with
multiple runs of the same experiment producing similar \emph{error
\%}.

Augmentation in data-space using elastic deformations gave a better
improvement in \emph{error \%} than augmentation in feature-space.
Interestingly, while \emph{test error \%} decreased, \emph{training
error \%} also decreased, and thus the gap between test and training
was roughly maintained. 

Augmentation in feature-space using SMOTE showed marginally promising
results. Using SMOTE, the \emph{test error \%} continued to decrease
as the number of samples were increased all the way to 50,000 samples.

Augmentation in feature-space using DBSMOTE resulted in a slight improvement
on test \emph{error \%}. Another interesting observation is that the
training error \% for SMOTE and DBSMOTE followed very similar curves.

\subsection{Convolutional SVM Results}

\begin{figure}[t]
\includegraphics[width=0.99\columnwidth]{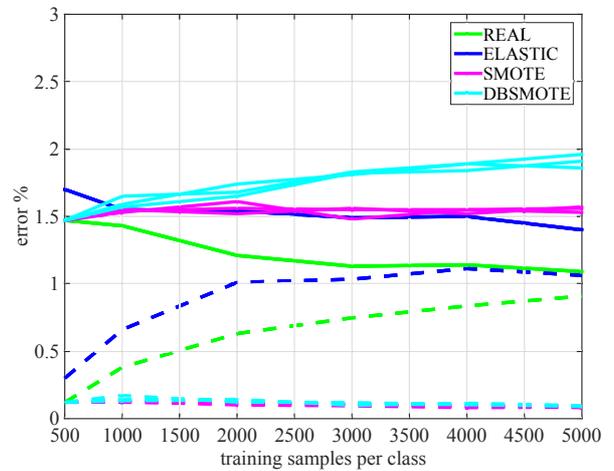}

\caption{CSVM performance on MNIST. The dashed lines indicate training error
\%, and the solid lines indicate test error \%.}
\label{figure_four}
\end{figure}

The performance of the CSVM is shown in Figure \ref{figure_four},
which illustrates how \emph{error \%} varies as the number of samples
are increased. Perhaps the most interesting result is that increasing
the number of synthetic samples using DBSMOTE caused the performance
to degrade, i.e., the \emph{error \%} increased as more samples were
used. Augmentation in feature-space using the DBSMOTE  algorithm was
not effective in reducing overfitting in the SVM classifier. However,
this result is not surprising as the DBSMOTE promotes towards the
centre of class clusters. By creating synthetic samples that are ``more
of the same'' DBSMOTE encourages overfitting of the training data.
Augmentation in data-space using the SMOTE algorithm provided little
to no improvement in performance. Even augmentation in data-space
using elastic warping only provided a modest improvement in \emph{test
error \%}, and this only occurred at very large amounts of data augmentation.
However, it should be noted that the gap between training \emph{error
\%} and \emph{test error \%} did decrease steadily as the amount of
augmentation was increased, thus indicating a reduction in overfitting.

\subsection{Convolutional ELM Results}

\begin{figure}[t]
\includegraphics[width=0.99\columnwidth]{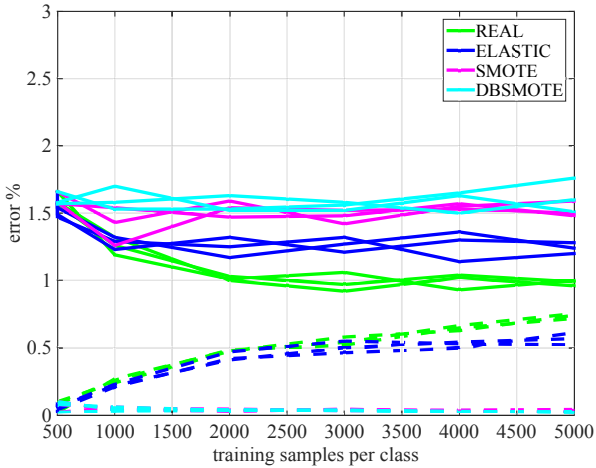}

\caption{CELM performance on MNIST. The dashed lines indicate training error
\%, and the solid lines indicate test error \%.}
\label{figure_five}
\end{figure}

The performance of the CELM is shown in Figure \ref{figure_five}.
This Figure illustrates how the \emph{training error} and \emph{test
error} varies as the number of samples are increased, which again
shows that performance typically improves as the amount of synthetic
samples are increased. However, this improvement is less than that
given by increasing the number of real samples. 

Augmentation in data-space using elastic deformations gave the best
results, which were slightly worse than having additional real training
samples. However the trend is not linear, as there was a marked 
improvement in \emph{error \%} for 1000 training samples
(500 real samples and 500 augmented samples) per class, before the gain in \emph{error
\%} flatten-off once again.

Augmentation in feature-space using SMOTE seemed promising when increasing
the amount of samples from 500 to 1000 samples per class. However,
increasing the amount of synthetic samples further resulted in a decrease
in performance. Also while \emph{test error \%} did initially decrease
with the use of SMOTE, there was no corresponding reduction in training
accuracy, which does occur when increasing the number of real samples.
Thus, the gap between training and testing performance remained large. 

Augmentation in feature-space using DBSMOTE had a slightly negative
impact on \emph{test error \%}. 

\section{Discussion}

The experiments conducted used a classification architecture that
was neatly divided into two stages: a feature generation stage and
a classification stage. The purpose of this was to investigate if
it is better to conduct data augmentation in \emph{data-space }or
\emph{feature-space}? 

For the experiments conducted on handwritten digits recognition, it
was clearly better to perform augmentation in \emph{data-space} using
elastic distortions. We expect this to hold true for other classification
tasks, if label preserving transforms are known and can be applied. 

When label preserving transforms are not avaliable, the SMOTE algorithm
can be used to perform data augmentation in \emph{feature-space},
and provide some benefit to a CNN or CELM classifier. Our results
suggest the DBSMOTE algorithm should not be used for data augmentation.
However, a classification architecture that is neatly divided into
\emph{data-space} and \emph{feature-space} is an artificial construct.
The modern trend is to construct architectures with deep layered hierarchies
of feature transformations with more complex features built upon simpler
features. Nevertheless, these results should also provide insight
into performing data augmentation for more modern architectures. A
good overview of label preserving transformations for image data for
a more modern classfication architecture is given by \cite{howard2013some}.

Another research question that we sought to answer, is how much data
is enough? For most classification systems, more data is better. And
more real data is better than more synthetic data. In none of the
experiments did the performance of the system trained on synthetic
data outperform the system trained on real data. Thus, the performance
(test error \%) that can be achieved by augmenting classifier training
with synthetic data, is likely to be bounded by training on the equivalment
amount of real data. 

For the CNN, adding more synthetic data, using SMOTE and elastic warping,
consistently reduced the testing error \%. The experiments did not
reach the limits of the amount of synthetic data that could be added
before error \% no longer improved.

For the CSVM adding more synthetic data, using DBSMOTE, caused classification
performance to degrade.

For the CELM more synthetic data was not always better for performance.
For the combination of the CELM with SMOTE there was a peak improvement
provided by data augmentation at 1000 training samples per class,
while increasing the number of samples further using the SMOTE technique
resulted in a decrease in performance.  

When comparing the three classification algorithms, the CSVM algorithm
demonstrated less benefit from data-augmentation than the CELM, which
showed less benefit than the CNN algorithm.

\section{Conclusion}

This paper demonstrates the benefits and pitfalls of data augmentation
in improving the performance of classification systems. Data augmentation
can be performed in data-space or feature-space. We found that it
was better to perform data augmentation in \emph{data-space}, as long
as label preserving transforms are known. The highly cited SMOTE algorithm
can be used to perform data augmentation in \emph{feature-space}.
This is a more robust solution than the DBSMOTE algorithm, which can
increase overfitting due to the algorithm creating new samples close
to existing cluster centers. The improvement in \emph{error \%} provided
by data augmentation was bounded by the equivalent amount of real-data.



%

%
%

\bibliographystyle{IEEEtran}
\bibliography{IEEEabrv,datawarping_references}

\end{document}